\documentclass[10pt,a4paper,conference]{IEEEtran}
\usepackage{tikz, graphicx}
\usepackage{comment}
\usepackage{amsmath,amssymb} 
\usepackage{color}
\usepackage{microtype}
\usepackage{makecell}
\usepackage{subfigure}
\usepackage{float}
\usepackage{booktabs}
\usepackage{amsfonts}
\usepackage{bm}
\usepackage{mathtools}
 \pdfoutput=1
\ifCLASSINFOpdf
  
\else
  
\fi

\begin{document}
\title{Fine-Tuning DARTS for Image Classification}
\author{\IEEEauthorblockN{Muhammad Suhaib Tanveer}
\IEEEauthorblockA{School of Electrical Engineering\\
Korea  Advanced  Institute  of \\ Science
and Technology (KAIST)\\ 
Daejeon, 34141 South Korea\\
Email: suhaibtanveer@kaist.ac.kr}
\and
\IEEEauthorblockN{Muhammad Umar Karim Khan}
\IEEEauthorblockA{Centre of Integrated Smart Sensors (CISS) \\
KAIST, Daejeon, 34141 South Korea\\}
\and
\IEEEauthorblockN{Chong-Min  Kyung}
\IEEEauthorblockA{School of Electrical Engineering\\
Korea  Advanced  Institute  of \\ Science
and Technology (KAIST)\\ 
Daejeon, 34141 South Korea\\}}
\maketitle

\begin{abstract}
Neural Architecture Search (NAS) has gained attraction due to superior classification performance. Differential Architecture Search (DARTS) is a computationally light method. To limit computational resources DARTS makes numerous approximations. These approximations result in inferior performance. We propose to fine-tune DARTS using fixed operations as they are independent of these approximations. Our method offers a good trade-off between the number of parameters and classification accuracy. Our approach improves the top-1 accuracy on Fashion-MNIST, CompCars, and MIO-TCD datasets by 0.56\%, 0.50\%, and 0.39\%, respectively compared to the state-of-the-art approaches.\ Our approach performs better than DARTS, improving the accuracy by 0.28\%, 1.64\%, 0.34\%, 4.5\%, and 3.27\% compared to DARTS, on CIFAR-10, CIFAR-100, Fashion-MNIST, CompCars, and MIO-TCD datasets, respectively.
\end{abstract}

\IEEEpeerreviewmaketitle

\section{Introduction}
\label{Introduction}
Image classification is a fundamental computer vision task. Image classification is employed in a number of different industries. Some popular industries that employ image classification are automobile, retail, security and healthcare industry.

Several approaches have been employed for image classification. Since the advent of deep learning, handcrafted deep neural networks are being used to achieve the state-of-the-art accuracy on the image classification task. Discovering state-of-the-art neural network architectures requires significant effort of human experts. Therefore, Neural Architecture Search (NAS), has been widely adopted, allowing automatic design of neural networks. NAS methods have achieved astonishing results on the image classification task, surpassing the manual methods on many popular datasets.

Various approaches have been employed for NAS. Non-Stochastic NAS methods are able to find good architectures but at a very high search cost. These methods use reinforcement learning \cite{zoph2018learning:1}, evolutionary algorithms \cite{RealAggarwalHuangLe2018:5} and sequential model-based optimization \cite{Liu_2018:4} techniques. On the other hand, stochastic methods are able to give competitive performance in a significantly shorter time.

DARTS \cite{liu2018darts:6} is a widely used stochastic method but it makes numerous approximations to speedup. Approximations such as using finite difference approximation and using weights of one forward step instead of the optimal weights lead to inferior performance.

Fine-tuning  is  a proven method  for  improving  performance of a neural network.
It is a process that uses an already trained neural network on a given task and makes it perform another downstream similar task. Inspired by fine-tuning, we propose incorporating fixed-operations to fine-tune DARTS. We use attention modules as fixed operations in our approach owing to the proven success of attention modules in improving classification accuracy \cite{hu2018squeeze:20}. Although our method performs automatic architecture search, incorporating manually-designed operations improves classification performance allowing us to search better architectures in the same amount of time compared to DARTS. 

The rest of the paper is structured as follows. Previous
work is detailed in Section 2. In Section 3, we provide detailed insights about our proposed method. Section 4 presents the experimental
results. The paper is concluded in Section 5.
\begin{figure*}
\includegraphics[width=\textwidth, height=7cm]{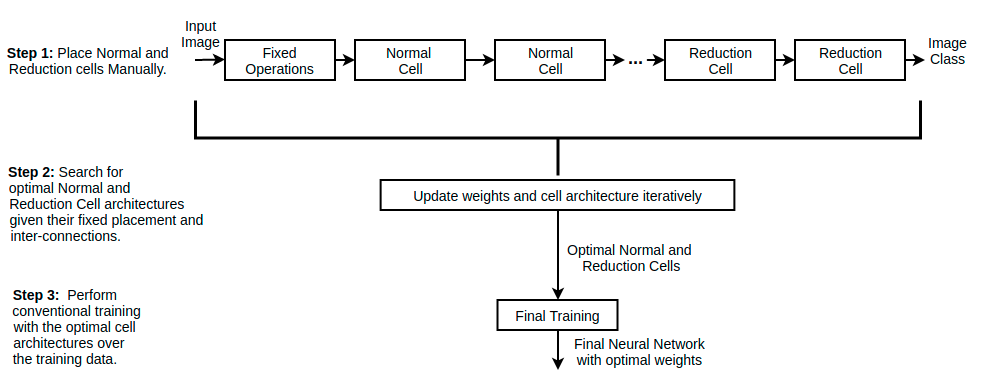}
\caption{Step-wise depiction of DARTS.}
\label{fig:13}
\end{figure*}
\section{Related Work}
Neural Architecture search (NAS) has been successfully applied to design model architectures for image classification. Reinforcement learning, evolutionary algorithms, sequential model-based and stochastic gradient-based optimization approaches are usually used for NAS.
\subsection {Non-Stochastic NAS Methods}
NAS-Net \cite{zoph2018learning:1} uses reinforcement learning with a controller RNN to design a cell. NAS-Net shows remarkable results but it requires a search cost of 2000 GPU days. Amoeba-Net \cite{RealAggarwalHuangLe2018:5} uses an evolutionary method to search for the optimal architecture. The results are superior compared to NAS-Net but search time is 3150 GPU days. Progressive Neural Architecture Search (PNAS) \cite{Liu_2018:4} uses a sequential model-based optimization strategy to guide the search through the search space. PNAS drastically reduces the search time to 225 GPU days while achieving competitive accuracy.
\subsection{Stochastic NAS methods}
 Neural Architecture Optimization Network (NAO-Net) \cite{luo2018neural:9} uses an encoder, predictor and a decoder. The encoder maps neural network architectures to a continuous space. The predictor uses that continuous representation as input and predicts the accuracy. The decoder maps that continuous representation back to the neural network architecture. NAO-Net reduces the search time to 200 GPU Days and produces good results.
 Self-Evaluated Template Network (SETN) \cite{dong2019one:12} proposes to use an evaluator and a template network. 
 SETN achieves impressive results at a very low search cost. SMASH \cite{brock2017smash:10} proposes to accelerate architecture selection by learning an auxiliary Hyper-Net. The auxiliary Hyper-Net generates the weights of the main model conditioned on that model’s architecture. By comparing the relative validation performance of networks with Hyper-Net-generated weights, they effectively search over a wide range of architectures at the cost of a single training run. 

In Stochastic Neural Architecture Search (SNAS) \cite{xie2018snas:4} the search space is represented with a set of one-hot random variables from a fully factorizable joint distribution, multiplied to mask operations in the graph. 
They reduce the search cost significantly while giving competitive performance. 

 Gradient-based search using Differentiable Architecture Sampler (GDAS) \cite{dong2019searching:13} develops a differentiable sampler over the Directed acyclic graph (DAG) to avoid traversing all the possibilities of the sub-graphs. 
 GDAS is quite fast but lacks accuracy.

\subsection{DARTS and its derivatives}
Differentiable Architecture Search (DARTS) \cite{liu2018darts:6} introduces a differentiable and continuous search space instead of a discrete search space and achieves remarkable efficiency, incurring a low search cost. Several methods have been proposed to improve DARTS. Progressive Differentiable Architecture Search (PDARTS) \cite{chen2019progressive:18} proposes to progressively increase the depth of the network during search while decreasing the search space to cater to the resource constraint. Partial Channel Connections for Memory-Efficient Differentiable Architecture Search (PC-DARTS) \cite{xu2019pc:17} leverages the redundancy in network space and samples a small portion of a super-net only via partial channel connections. 

Prune and Replace DARTS (PR-DARTS) \cite{laube2019prune:14} uses a small candidate operation pool from which candidates are progressively pruned and replaced with better performing ones. Amended-DARTS \cite{bi2019stabilizing:16} proposes an amending term for computing architectural gradients by making use of a direct property of the optimality of network parameter optimization. DARTS+ \cite{liang2019darts+:19} proposes to use an early stopping criterion to improve efficiency.
\nobreak
\begin{figure}[ht]
\vskip 0.2in
\begin{center}
\centerline{\includegraphics[width=1.08\linewidth, height=18cm]{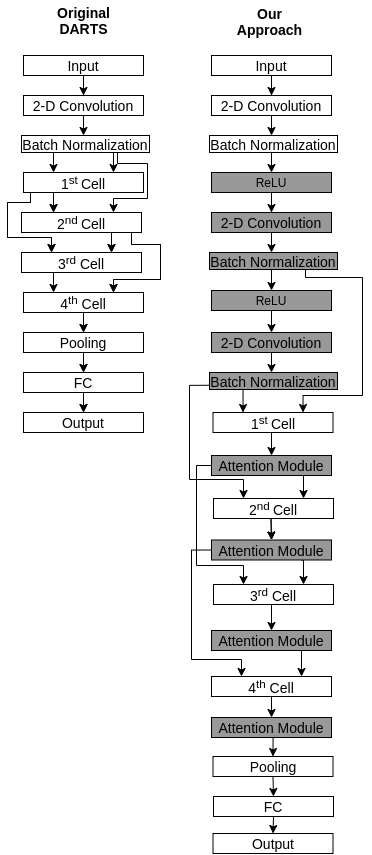}}
\caption{Comparison of DARTS \cite{liu2018darts:6} and our approach for a four cell network. Block
marked in grey indicate the difference between the two approaches.}
\label{1}
\end{center}
\vskip -0.2in
\end{figure}

\begin{figure}
\hskip 0.6in
\includegraphics[width=0.25\textwidth,height=13.5cm]{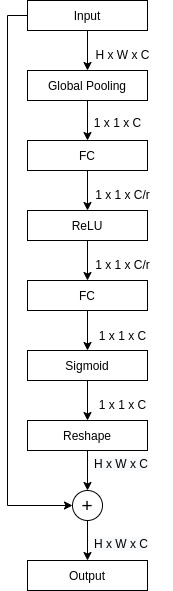}
\caption{The schema of attention module. 'FC' represents fully-connected layer.}
\label{21}
\vskip -0.2in
\end{figure}
\section{Fine-tuning DARTS}
DARTS proposes a method where
the architecture is updated by minimizing the validation loss using the delta rule. In other words, the neural architecture itself is made differentiable with the validation loss.
In DARTS, a predefined number of cells are stacked
together to form the neural network. A cell is a directed
acyclic graph (DAG) consisting of a pre-defined number of
nodes. A node is a feature map in convolution networks.
Each cell has seven nodes, two input nodes, four intermediate
nodes and one output node. Input nodes are the output nodes
of previous two cells. Intermediate nodes have two inputs and one output. There are two types of cells: normal
and reduction. The reduction cells reduce the spatial resolution of the features by half whereas the normal cells leave
the feature resolution unchanged. The process is shown in Fig. \ref{fig:13}. The features in a cell are represented by nodes. The number of nodes are fixed from the beginning. 

\subsection{Approximations in DARTS}
Multiple approximations have been used in DARTS, leading to reduced performance. First, the architecture $\alpha$ is updated after a single update of the weights $w$ rather than the optimal weights $w *(\alpha)$. More specifically
\begin{align}
&\nabla_\alpha \mathcal{L}_{val}(w^*(\alpha), \alpha) \\
\approx &\nabla_\alpha \mathcal{L}_{val}(w - \xi \nabla_{w} \mathcal{L}_{train}(w, \alpha), \alpha), \label{eq:single}
\end{align}
where $\mathcal{L}_{train}$, $\mathcal{L}_{val}$ and $\xi$ represent the training loss, validation loss and learning rate for weight update, respectively.
Second, a finite difference approximation has been used to reduce the computational complexity of obtaining the gradient of the architecture $\alpha$ with the validation data. With these approximations, the convergence of DARTS to the optimal architecture is yet to be theoretically proven. These approximation are crucial to DARTS as these provide the significant speedup. However, as common to approximations, these are expected to lead to inferior performance.
​\subsection{Proposed Solution}
Fine-tuning is a time-tested method for improving performance of a neural network. The parameters of a neural network pre-trained with a different dataset are adopted for the current task. In other words, few or all parameters of the neural network are initialized with the parameters of a pre-trained network and trained over the current data. Fine-tuning provides the easiest way of transferring information across datasets, thereby, allowing neural networks to learn quickly as well as improve performance. Different variants of fine-tuning have been adopted where some layers of the neural network are frozen to the pre-trained values while training.

Before the recent advancements of NAS methods, manually designed neural architectures have achieved a great degree of success. In the seminal work of AlexNet \cite{krizhevsky2012imagenet:55}, the authors proposed a neural network, which significantly outperformed its predecessors. Subsequently, the method was further improved. ResNet \cite{7780459:61} introduces residual blocks with skip connections, which allow the gradients to back propagate through to the initial layers without vanishing. More recently, Attention Modules \cite{hu2018squeeze:20} have been introduced, which allow the neural networks to focus on regions of interest before making a decision.
​

NAS is an interesting theoretical problem as it introduces algorithms to obtain neural architecture from scratch. However, NAS methods do not make use of the years of effort towards manually designing neural architecture. Our idea is to fine-tune neural architecture search to obtain better performance. By this we utilize manually-designed architectures in the DARTS method.
​

In original DARTS, there are $M$ possible connections and $N$ possible operations. Thus, the cell architecture represented by a matrix $\alpha \in \mathbb{R}^{M \times N}$. We propose to extend $\alpha$ as $\bm{\alpha} \in \mathbb{R}^{M'\times N'}$, where $M' > M$ and $N' \ge N$. Mathematically, this extension can be represented as
​
\begin{equation}
    \bm{\alpha} = [\alpha; \alpha_F],
\end{equation}where $\alpha_F$ represents fixed operations and $;$ represents concatenation. Columns of zeros are included if $N' > N$. The value of $N'$ is determined as
​
\begin{equation}
    N' = N + |O(\alpha^F) - O(\alpha)|,
\end{equation}where $O(\alpha)$ is the set of operations of $\alpha$ and $|.|$ represents the cardinality of the set. Note that $N' = N$ if all the operations in $\alpha_F$ exist in $\alpha$.
When the architecture $\bm{\alpha}$ is updated, only the architecture parameters from $\alpha$ are updated and $\alpha_F$ remains fixed. More specifically
​
\begin{equation}
    \bm{\alpha}^{(i,j)} \coloneqq 
        \begin{cases}
            \bm{\alpha}^{(i,j)} - \gamma \Delta_{\bm{\alpha}}^{(i,j)}\mathcal{L}_{val}(w', \bm{\alpha}) & \text{if } i \le M, j \le N \\
            \bm{\alpha}^{(i,j)} & \text{otherwise},
        \end{cases}
\end{equation}where $\gamma$ is the learning rate for architecture update, $i\in \{1, 2, ..., M'\}$, $j \in {1, 2, ..., N'}$ and 
​
\begin{equation}
    w' = w - \xi \Delta_w\mathcal{L}_{train}(w, \bm{\alpha}).
\end{equation}

By introducing $\alpha_F$, we expect two distinct advantages. First, $\alpha_F$ can be based on any recent manually-designed architecture, thus making use of past efforts in architecture design. Second, part of the architecture is somewhat independent of the DARTS method. In other words, the approximations of the DARTS method do not carry their influence to $\alpha_F$ resulting in robust performance. In Fig. \ref{1} we show the difference between our approach and original DARTS.

\subsection{Architecture}
 We use the attention modules described in \cite{hu2018squeeze:20} as $\alpha_F$ as it has been recently proposed, highly intuitive and shows good performance. The attention module adaptively recalibrates channel-wise feature responses by explicitly modeling interdependencies between channels. 
 
 
 Fig. \ref{21} depicts the architecture of our attention module. In the squeeze phase, each attention module makes use of a global average pooling operation. The squeeze phase is followed by the excitation phase which makes use of two small fully-connected layers. The excitation phase is followed by an inexpensive channel-wise scaling/reshaping operation. Reduction ratio $r$ is a hyperparameter that allows us to vary the capacity and computational cost of the attention module in the network. Following \cite{hu2018squeeze:20} we set the reduction ratio $r$ as 16 in our experiments. 
  Attention module has 2 nodes (3 connections) and 3 operations, so $\bm{\alpha} \in \mathbb{R}^{17 \times 11}$.
 
\begin{figure}
\begin{center}
\includegraphics[width=0.5\textwidth, height=11cm]{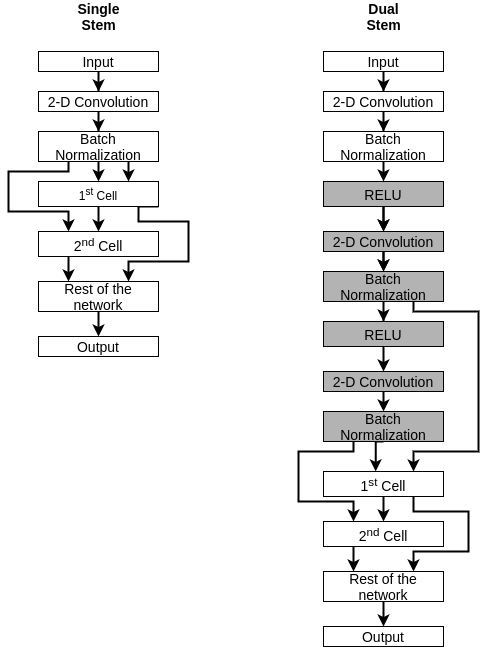}
\caption{Comparison of single-stem and dual-stem approach. Blocks marked in gray indicate the difference between the two approaches.}
\label{fig:7}
\end{center}
\end{figure}
\begin{figure}
\begin{center}
\includegraphics[width=0.5\textwidth, height=3.5cm]{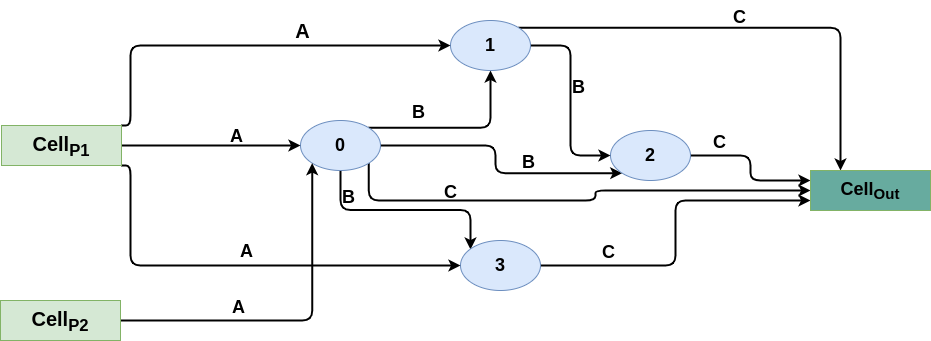}
\caption{Reduction cell found on Fashion-MNIST dataset using DARTS. \textit{$Cell_{Out}$} represents the output node of the cell, \textit{$Cell_{P1}$} and \textit{$Cell_{P2}$} represent outputs of previous two cells respectively, and 0, 1, 2 and 3 are intermediate nodes of the cell. \textbf{A}: Average Pooling $3\times3$, \textbf{B}: Skip Connection and \textbf{C}: Concatenation.}
\label{fig:4}
\end{center}
\end{figure}

\begin{figure}
\begin{center}
\includegraphics[width=0.50\textwidth, height=3.5cm]{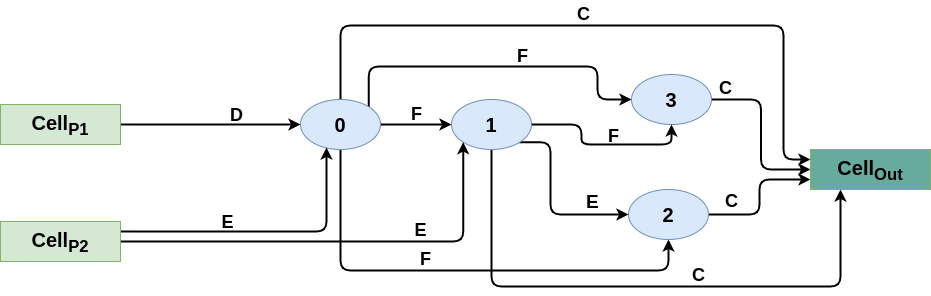}
\caption{Reduction cell found on Fashion-MNIST dataset using our method. \textit{$Cell_{Out}$} represents the output node of the cell, \textit{$Cell_{P1}$} and \textit{$Cell_{P2}$} represent outputs of previous two cells respectively, and 0, 1, 2 and 3 are intermediate nodes of the cell. \textbf{C}: Concatenation, \textbf{D}: Separable Convolution $5\times5$, \textbf{E}:  Dilated Separable Convolution $5\times5$ and \textbf{F}: Maximum Pooling $3\times3$.}
\label{fig:5}
\end{center}
\end{figure}

\floatstyle{plaintop}
\restylefloat{table}
\begin{table*}[tbp]
    \centering
    \begin{tabular}{c|c|c|c|ccc}
    \toprule
    Architecture & Accuracy(\%) & Params(M) & \makecell{Search \\ Method} & \makecell{Search Cost \\ GPU days}\\
    \midrule
   DenseNet-BC \cite{huang2017densely:25} & 96.54 & 25.6 & Manual & -\\
   \hline
NASNetA + cutout \cite{zoph2018learning:1} & 97.35 & 3.3 & RL & 2000\\
AmoebaNet-B + cutout \cite{RealAggarwalHuangLe2018:5} & \textbf{97.87} & 34.9 & Evolution & 3150\\
PNAS \cite{Liu_2018:4} & 96.59 $\pm$ 0.09 & 3.2 & SMBO & 225\\
NAONet \cite{luo2018neural:9}  & 96.82 & 10.6 & NAO & 200\\
\hline
SMASHv2 \cite{brock2017smash:10} & 95.97 & 16 & GB & 1.5 \\
SETN + cutout \cite{dong2019one:12} & 97.31 & 4.6 & GB & 1.8\\
GDAS + cutout \cite{dong2019searching:13}  & 96.25 & 2.5 & GB & 0.17\\
DARTS(2nd order) + cutout \cite{liu2018darts:6} & 97.24 $\pm$ 0.09 & 3.3 & GB & 1\\
SNAS (mild) + cutout \cite{xie2018snas:4}& 97.02 & 2.9 & GB & 1.5\\
PR-DARTS DL1 + cutout \cite{laube2019prune:14} & 97.26 $\pm$ 0.12 & 3.2 & GB & 0.82\\
PC-DARTS +cutout \cite{xu2019pc:17} & 97.43  & 3.6 & GB & 0.1\\
PDARTS + cutout \cite{chen2019progressive:18} & 97.50  & 3.4 & GB & 0.3\\
Amended-DARTS S1 + cutout \cite{bi2019stabilizing:16} & 97.19 $\pm$ 0.21 & 3.5 & GB & 1.0 \\
DARTS+ with cutout \cite{liang2019darts+:19}   & 97.68 & 3.7 & GB & 0.4 \\
\hline
\textbf{Ours + cutout}  & 97.52 & 3.9 & GB & 1 \\
    \bottomrule
    \end{tabular}
    \caption{Top-1 Accuracy on CIFAR-10 dataset. GB:Gradient-Based}
    \label{tab:cifar10}
\end{table*}
\begin{table*}[tbp]
    \centering
    \begin{tabular}{c|c|c|c|ccc}
    \toprule
    Architecture & Accuracy (\%) & Params(M) & \makecell{Search \\ Method} &  \makecell{*Search Cost \\ GPU days}\\
    \midrule
   DenseNet-BC \cite{huang2017densely:25} & 82.82 & 25.6 & Manual & -\\
   \hline
NASNetA + cutout\ \cite{zoph2018learning:1} & 82.19 & 3.3 & RL & 2000\\
AmoebaNet-B + cutout \cite{RealAggarwalHuangLe2018:5} & 84.20 & 34.9 & Evolution & 3150\\
PNAS  + cutout \cite{Liu_2018:4}& 82.37 & 3.2 & SMBO & 225\\
NAONet \cite{luo2018neural:9}  & \textbf{84.33} & 10.6 & NAO & 200\\
\hline
SETN + cutout \cite{dong2019one:12} & 82.75 & 4.6 & Gradient-based & 1.8\\
GDAS + cutout \cite{dong2019searching:13}  & 80.91 & 2.5 & Gradient-based & 0.17\\
\makecell{DARTS(2nd order) \\ + cutout \cite{liu2018darts:6}} & 82.46 & 3.3 & Gradient-based & 1\\
PDARTS + cutout \cite{chen2019progressive:18} & 82.80  & 3.4 & Gradient-based & 0.3\\
DARTS+ with cutout \cite{liang2019darts+:19}   & 83.72 & 3.7 & Gradient-based & 0.4 \\
\hline
\textbf{Ours + cutout}  & 84.10 & 3.9 & Gradient-based & 1 \\
    \bottomrule
    \end{tabular}
    \caption{Top-1 Accuracy on CIFAR-100 dataset. *: Search cost on CIFAR-10.}
    \label{tab:cifar100}
\end{table*}
\begin{table*}[tbp]
    \centering
    \begin{tabular}{c|c|c|ccc}
    \toprule
    Architecture & Accuracy (\%) & Params(M) & \makecell{Search \\ Method} \\
    \midrule
   ResNet-110 + random erasing \cite{zhong2017random:27} & 95.99 $\pm$ 0.13 & 1.7 & Manual \\
    ResNeXt-8-64 + random erasing \cite{zhong2017random:27} & 96.21 $\pm$ 0.06 & 34.4 & Manual \\
   WRN-28-10 + random erasing \cite{zhong2017random:27} & 96.35 $\pm$ 0.03 & 36.5 & Manual \\ 
   VGG8B \cite{nokland2019training:31} & 95.47 & 7.3 & Manual \\
   DeepCaps \cite{Rajasegaran_2019_CVPR:34} & 94.46  & 7.2 &Manual \\
   Neupde \cite{sun2019neupde:32} & 92.40 & 0.4 & Manual \\
\hline
DARTS(2nd order) + cutout + random erasing \cite{liu2018darts:6} & 96.57 & 2.6 & GB\\
\hline
\textbf{Ours + cutout + random erasing}  &\textbf{96.91} & 3.2 & GB\\
    \bottomrule
    \end{tabular}
    \caption{Top-1 Accuracy on Fashion-MNIST dataset. GB:Gradient-Based.}
    \label{tab:mnist}
\end{table*}
\begin{table}
\begin{center}
\begin{tabular}{|l|c|l|c|}
\hline
Method & Nature of Data  & CNN Type & Accuracy \\
\hline\hline
AlexNet{\cite{DBLP:journals/corr/YangLLT15:3}} & Web & Manual & 81.9\% \\
Overfeat{\cite{DBLP:journals/corr/YangLLT15:3}} & Web & Manual & 87.9\% \\
GoogLeNet{\cite{DBLP:journals/corr/YangLLT15:3}} & Web & Manual & 91.2\% \\
Han et al.\cite{Han:2018:AAM:3240508.3240550:28} & Web & Manual & 95.4\% \\
DARTS \cite{liu2018darts:6} &  Web & NAS-based & 91.4\%\\
SNAS \cite{xie2018snas:4}& Web & NAS-based& 91.4\%\\
Ours & Web & NAS-based & \textbf{95.9\%} \\
\hline
AlexNet{\cite{DBLP:journals/corr/YangLLT15:3}} & Surveillance & Manual & 98.0\% \\
Overfeat{\cite{DBLP:journals/corr/YangLLT15:3}} & Surveillance & Manual & 98.3\% \\
GoogLeNet{\cite{DBLP:journals/corr/YangLLT15:3}} & Surveillance & Manual & 98.4\% \\
Fang et al.\cite{Fang2017FineGrainedVM:27} & Surveillance & Manual & 98.6\% \\
DARTS \cite{liu2018darts:6} & Surveillance & NAS-based & 98.1\%\\
SNAS \cite{xie2018snas:4}& Surveillance & NAS-based& 98.4\%\\
Ours & Surveillance & NAS-based & \textbf{99.2\%} \\ 
\hline
\end{tabular}
\end{center}
\caption{Top-1 Accuracy on CompCars dataset.}
\label{tab:1}
\end{table}
\begin{table}
\begin{center}
\begin{tabular}{|l|c|l|}
\hline
Method & CNN Type & Accuracy \\
\hline\hline
Xception\cite{chollet2017xception:44} & Manual & 97.61\% \\
Theagarajan et al.\cite{theagarajan2017eden:45} & Manual & 97.80\% \\
Kim et al.\cite{kim2017vehicle:46} & Manual & 97.86\% \\
Lee et al.\cite{taek2017deep:47}& Manual & 97.92\% \\
Jung et al.\cite{jung2017resnet:48} & Manual & 97.95\% \\
DARTS \cite{liu2018darts:6} & NAS-based & 95.07\% \\ 
SNAS \cite{xie2018snas:4} & NAS-based & 95.50\% \\
Ours & NAS-based & \textbf{98.34\%} \\
\hline
\end{tabular}
\end{center}
\caption{Top-1 Accuracy on MIO-TCD dataset.}
\label{tab:3}
\end{table}
\section{Experimental results}
In this section, we discuss applying our method to numerous public image classification datasets. For our experiments, we used an Nvidia GTX 1080 GPU with 8GB of memory, which is a relatively small GPU. The computer used for running the experiments had an Intel(R) Core(TM) i7-3770K CPU and 8GB of RAM. 
  For experiments on CompCars and MIO-TCD we use a dual-stem approach instead of a one-stem approach because these datasets have images of size larger than 32$\times$32. Rather than passing redundant information to the first and the second cells of the neural network, we propose applying two different transformations to the input image, and passing the first transformed image to the first cell and the second transformed image to the second cell. In this way, different information is passed to the first two cells of the neural network. A comparison of both approaches is shown in Fig. \ref{fig:7}. Results show that our dual-stem approach works better on these datasets compared to single-stem approach of DARTS. 
\subsection{CIFAR-10 Dataset}
CIFAR-10 \cite{krizhevsky2009learning:21} is a dataset for image classification. The CIFAR-10 dataset consists of 60,000 $32 \times 32$ color images divided in 10 classes, with 6000 images per class. There are 50000 training images and 10000 test images. 

For the CIFAR-10 dataset, we used an initial learning rate of 0.025 and the learning rate was updated using the strategy in \cite{Loshchilov2017SGDRSG:22} to learn the network parameters. The momentum and weight decay parameters of stochastic gradient descent (SGD) optimizer were set to 0.9 and $3\times 10^{-4}$. For the cell search, we used $3\times 10^{-4}$, 0.5, 0.999 and $10^{-3}$ as the learning rate, $\beta_1$, $\beta_2$ and weight decay values of Adam optimizer, respectively. 50\% of the training data was used as validation data during the architecture search. For the final training, the standard training/testing split is used.We stacked 8 and 20 cells during architecture search and final training after architecture search respectively. A batch size of 32 and 56 were used during architecture search and final training, respectively. The initial number of channels during architecture search and final training were set to 16 and 36, respectively. We performed cell search and final training of the neural network over 50 and 600 epochs of the training data, respectively. Additional enhancements during final training include cutout \cite{devries2017improved:24}, path dropout of probability 0.2 and auxiliary towers with weight 0.4.

We give a detailed comparison of our approach with all the approaches mentioned in Section 2 in Table \ref{tab:cifar10}. Although AmoebaNet \cite{real2019regularized:7}, gives better accuracy but the number of parameters and search cost are too high. NAONet \cite{luo2018neural:9}, gives the same accuracy as compared to our method but the number of parameters and search cost is higher. Although some of the methods ( GDAS, PR-DARTS, PC-DARTS, PDARTS, and DARTS+ ) search faster compared to our method. However, the GPU days metric depends upon the type of the GPU used, as some of these methods use a different GPU so it's not a very accurate metric. 
\subsection{CIFAR-100 Dataset}
\noindent CIFAR-100 \cite{krizhevsky2009learning:21}, is a large-scale dataset for image classification. The CIFAR-100 dataset consists of 60,000 $32 \times 32$ color images in 100 classes, with 600 images per class. There are 500 training images and 100 test images per class.

We directly apply our best searched architecture from CIFAR-10 experiments to CIFAR-100. We follow the same training settings as we did for the final training stage of CIFAR-10. All the architecture search methods reported in Table \ref{tab:cifar100} use the architecture searched on the CIFAR-10 dataset to provide a fair comparison. Although NAONet \cite{luo2018neural:9} and AmoebaNet \cite{RealAggarwalHuangLe2018:5}, give more accuracy but the number of parameters are quite high compared to our method.

\subsection{Fashion-MNIST Dataset}
Fashion-MNIST \cite{xiao2017/online:26} is a dataset consisting of images related to clothe-ware, shoes, and bag.  The Fashion-MNIST dataset has a training set of 60,000 examples and a test set of 10,000 examples. Each example is a 28x28 gray-scale image, associated with a label from 10 classes. 

We used an initial learning rate of 0.025 and the learning rate was updated using the strategy in \cite{Loshchilov2017SGDRSG:22} to learn the network parameters. The momentum and weight decay parameters of stochastic gradient descent (SGD) optimizer were set to 0.9 and $3\times 10^{-4}$. For the cell search, we used $3\times 10^{-4}$, 0.5, 0.999 and $10^{-3}$ as the learning rate, $\beta_1$, $\beta_2$ and weight decay values of Adam optimizer, respectively. 40\% of the training data was used as validation data during the architecture search. 15\% of the training data was used as validation data during final training. We stacked 8 and 20 cells during the architecture search and final training after architecture search respectively. A batch size of 32 and 72 were used during architecture search and final training, respectively. The initial number of channels during architecture search and final training were set to 16 and 36, respectively. We performed cell search and final training of the neural network over 50 and 600 epochs of the training data, respectively. Additional enhancements during final training include cutout \cite{devries2017improved:24}, path dropout of probability 0.2 and random erasing \cite{zhong2017random:27}.

We give a comparison of our approach with other state-of-the-art approaches on Fashion-MNIST dataset in Table \ref{tab:mnist}.
For a fair comparison, we use similar training settings to report results of DARTS on the Fashion-MNIST dataset. In accordance with our theory, our method performs better compared to DARTS.
Although the model proposed by Neupde \cite{sun2019neupde:32} uses only 0.4M parameters, its error rate is higher compared to other methods. Results clearly show the superiority of our method compared to other approaches as we are able to achieve state-of-the-art accuracy on Fashion-MNIST dataset. The reduction cell learned on the Fashion-MNIST dataset using DARTS \cite{liu2018darts:6} is shown in Fig. \ref{fig:4}. Reduction cell learned on Fashion-MNIST dataset using our method is shown in Fig. \ref{fig:5}. 

\subsection{CompCars Dataset}
\noindent CompCars \cite{DBLP:journals/corr/YangLLT15:3} is a large-scale dataset for fine-grained vehicle classification. CompCars dataset is further divided into two groups. The first group contains images of cars taken from the internet while the second group contains images taken from surveillance cameras. These images are challenging as they were taken in different weather and illumination conditions. The web-natured subset has 431 car models, with 35,456 and 15,627 images for training and testing, respectively. The surveillance-natured subset has 281 car models, with 31,146 and 13,333 training and testing images, respectively. 

For the CompCars dataset, we used an initial learning rate of 0.1 and the learning rate was updated using the strategy in \cite{Loshchilov2017SGDRSG:22} to learn the network parameters. The momentum and weight decay parameters of stochastic gradient descent (SGD) optimizer were set to 0.9 and $3\times 10^{-4}$. For the cell search, we used $3\times 10^{-4}$, 0.5, 0.999 and $10^{-3}$ as the learning rate, $\beta_1$, $\beta_2$ and weight decay values of Adam optimizer, respectively. Training images were further divided into training and validation subsets with a ratio of 60:40 and 70:30 respectively during architecture search and final training for web-nature data . 40\% of the training data was used as validation data during the architecture search. For the final training, we used 30\% (web-natured) and 20\% (surveillance-natured) of the training data for validation. We stacked 6 and 16 cells during architecture search and final training after architecture search respectively. A batch size of 32 and 72 were used during architecture search and final training, respectively. The initial number of channels during architecture search and final training were set to 16 and 48, respectively. For architecture search, we performed 80 and 50 epochs over the training data for the web and surveillance subsets, respectively whereas we performed 250 epochs for final training with both the subsets.

In Table \ref{tab:1}, we compare our approach  with DARTS\cite{liu2018darts:6}, SNAS \cite{xie2018snas:4}, manually designed CNNs that include AlexNet \cite{DBLP:journals/corr/YangLLT15:3} , Overfeat \cite{DBLP:journals/corr/YangLLT15:3}, GoogLeNet \cite{DBLP:journals/corr/YangLLT15:3}, and with state-of-the-art approaches on the web (Han \textit{et al.} \cite{Han:2018:AAM:3240508.3240550:28}) and surveillance subsets \cite{Fang2017FineGrainedVM:27}. As evident from the table, our approach gives the best accuracy compared to all other competing methods.

\subsection{MIO-TCD}
MIO-vision Traffic Camera Dataset (MIO-TCD) \cite{8387876:4} is the largest dataset for motorized-traffic analysis to date.  It is a challenging dataset because of the diversity of pose, lighting, inter-class similarity, and image resolution. MIO-TCD dataset contains 11 traffic object classes such as single-unit-truck, pickup-truck, and articulated-truck. The classification dataset consists of 648,959 images. The dataset is split into 80\% training (519,164) and 20\% testing(129,795) images.

We used the same values of $\beta_1$, $\beta_2$, momentum and the learning rate strategies as for the CompCars dataset. An initial learning rate of 0.025 was used. A batch size of 32 and 72 were used during architecture search and final training, respectively. The initial number of channels during architecture search and final training were set to 16 and 36, respectively. 35\% of the training data was used as validation data during the architecture search. For the final training, we used 20\% of the training data for validation. Again for time efficiency, we stacked 6 cells at architecture search and 14 cells during final training.

In Table \ref{tab:3}, we compare the top-1 accuracy on the MIO-TCD \cite{8387876:4} dataset of our approach with DARTS\cite{liu2018darts:6}, SNAS\cite{xie2018snas:4}, and with manual state-of-the-art approaches, which include \cite{chollet2017xception:44}, \cite{theagarajan2017eden:45}, \cite{kim2017vehicle:46}, \cite{taek2017deep:47}, and \cite{jung2017resnet:48}. Results show that our approach outperforms all other competing approaches.

\section{Conclusion}
Based on the proven success of fine-tuning in manually designed architectures we propose to fine-tune DARTS by adding fixed operations. We add attention modules after each cell. These operations are independent of the approximations used in DARTS. We conduct experiments on CIFAR-10, CIFAR-100, Fashion-MNIST, CompCars and MIO-TCD, and our results show the validity of our claim. We were able to obtain state-of-the-art results on Fashion-MNIST, CompCars and MIO-TCD datasets while our results on other datasets were also competitive.
   
\small{}
\bibliography{main}
\bibliographystyle{ieeetr}

\end{document}